\documentclass[conference]{IEEEtran}
\pdfoutput=1
\IEEEoverridecommandlockouts
\usepackage{cite}
\usepackage{amsmath,amssymb,amsfonts}
\usepackage{algorithmic}[1]
\usepackage{algorithm}
\newtheorem{theorem}{Theorem}
\usepackage{graphicx}
\usepackage{textcomp}
\usepackage{xcolor}
\usepackage{multirow}
\usepackage{booktabs}

\usepackage[colorlinks,linkcolor=red, anchorcolor=blue, citecolor=green]{hyperref}

\def\BibTeX{{\rm B\kern-.05em{\sc i\kern-.025em b}\kern-.08em
    T\kern-.1667em\lower.7ex\hbox{E}\kern-.125emX}}
\begin{document}

\title{Progressive Supervision via Label Decomposition: An Long-Term and Large-Scale Wireless Traffic Forecasting Method
}

\author{\IEEEauthorblockN{1\textsuperscript{st} Daojun Liang}
\IEEEauthorblockA{\textit{School of Information Science and Engineering} \\
\textit{Shandong University}\\
Qingdao 266237, China \\
liangdaojun@mail.sdu.edu.cn}
\and
\IEEEauthorblockN{2\textsuperscript{nd} Haixia Zhang}
\IEEEauthorblockA{\textit{School of Control Science and Engineering} \\
\textit{Shandong University}\\
Jinan, 250061, China \\
haixia.zhang@sdu.edu.cn}
\and
\IEEEauthorblockN{3\textsuperscript{rd} Dongfeng Yuan}
\IEEEauthorblockA{\textit{School of Qilu Transportation} \\
\textit{Shandong University}\\
Jinan, 250002, China \\
dfyuan@sdu.edu.cn}
}

\maketitle

\begin{abstract}
    Long-term and Large-scale Wireless Traffic Forecasting (LL-WTF) is pivotal for strategic network management and comprehensive planning on a macro scale. However, LL-WTF poses greater challenges than short-term ones due to the pronounced non-stationarity of extended wireless traffic and the vast number of nodes distributed at the city scale. To cope with this, we propose a Progressive Supervision method based on Label Decomposition (PSLD). 
    Specifically, we first introduce a Random Subgraph Sampling (RSS) strategy designed to sample a tractable subset from large-scale traffic data, thereby enabling efficient network training. Then, PSLD employs label decomposition to obtain multiple easy-to-learn components, which are learned progressively at shallow layers and combined at deep layers to effectively cope with the non-stationary problem raised by LL-WTF tasks.
    Finally, we compare the proposed method with various state-of-the-art (SOTA) methods on three large-scale WT datasets.
    Extensive experimental results show the proposed PSLD outperforms existing methods in terms of performance and runtime, with an average 2\%, 4\%, and 11\% performance improvement on three WT datasets, respectively. In addition, we built an open source library for WT forecasting (WTFlib) to facilitate related research, which contains numerous SOTA methods and provides a strong benchmark.
    Experiments can be reproduced through this anonymous link https://github.com/Anoise/WTFlib.
\end{abstract}


\section{Introduction}
\label{sec_intro}

Wireless networks have become an essential part of our daily lives, and the demand of wireless services is growing rapidly. 
For example, global mobile devices grow from 8.8 billion in 2018 to 13.1 billion in 2023, of which 1.4 billion is 5G capable \cite{cisco2020cisco}. 
The exponential growth of Wireless Traffic (WT) poses significant challenges, including the serious energy consumption of base stations \cite{narayanan2021variegated} and the inability to meet differentiated Quality of Service (QoS) requirements \cite{asghar2022evolution}.
WT forecasting can guide base stations to achieve power control \cite{niu2010cell} and flexible coverage \cite{kato2016deep, DeepCog2019}, which is an important means to achieve green and intelligent networks \cite{li2017intelligent}. 

\begin{figure}
    \centerline{\includegraphics[width=\columnwidth]{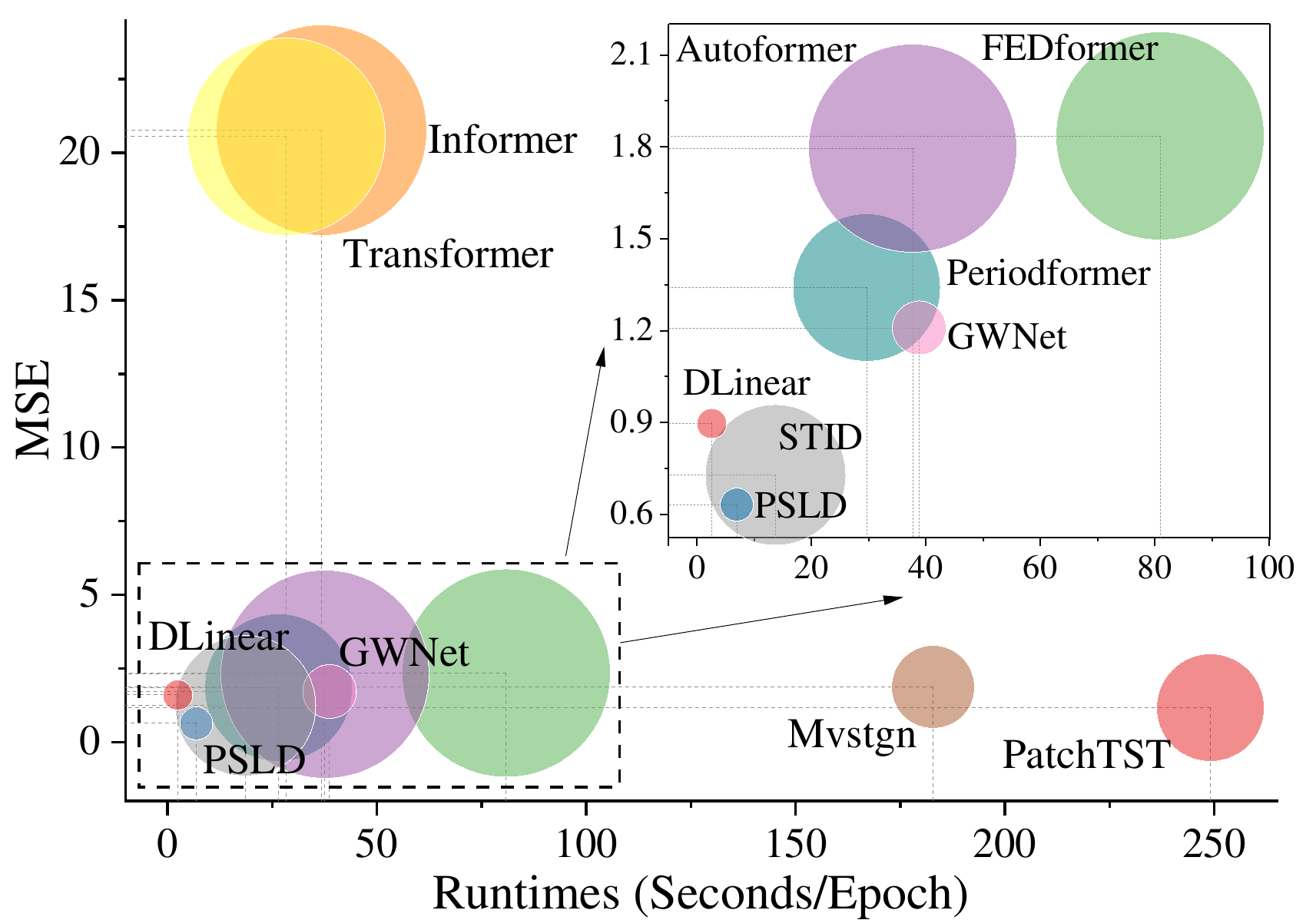}}
    \caption{Performance (MSE), running time (Seconds/Epoch) and Flops (Bubble Size) comparisons of time series models on the Milano dataset. The input lengths are both 36, and their prediction lengths are also 36. All experiments were performed on the Milano dataset using a Tesla V100 GPU. The smaller the bubble and the closer it is to the bottom left corner, the better the overall performance of the model will be better.}
    \label{fig1} 
\end{figure}


Traditional WT forecasting methods such as ARIMA \cite{shu2005wireless, zhou2006traffic, xu2016big}, entropy theory \cite{li2014prediction}, covariance function \cite{chen2015analyzing} and $\alpha$-stable distribution \cite{li2017learning} are developed specifically for short-term, low-dimensional time series. 
These methods are based on the stationarity assumption or statistical properties of time series to predict the future values, which are no longer suitable for high-dimensional and non-stationary time series existing in WT forecasting task \cite{de200625}. 
Recently, deep learning methods \cite{huang2017study, STDenseNet2018, STCNet2019, yao2021mvstgn, li2023dynamic} are developed for WT forecasting due to its powerful nonlinear fitting capabilities \cite{hornik1991approximation}. 
These methods either utilize Convolutional Neural Networks (CNNs) \cite{lecun1998gradient} and Graph Neural Networks (GNNs) \cite{scarselli2008graph} to process short-term grid data \cite{huang2017study, STDenseNet2018, STCNet2019} or Recurrent Neural Networks (RNNs) \cite{connor1994recurrent} to perform univariate prediction \cite{JASC_Meta2020, Forecasting2021, ICC_C2LM2021, hachemi2021mobile}.

However, there are two major problems that need to be addressed: 
1) Short-term WT forecasting (e.g., one-step-ahead prediction) cannot be used for long-term planning and lead to frequent decision changes, which in turn cause greater control overhead for the base station. 
Merely elongating the output span of current methodologies fails to address the inherent non-stationarity issue within long-term WT forecasting, thereby culminating in suboptimal predictive accuracy.
2) Wireless nodes are densely distributed and present a graph-like structure, making it impossible to efficiently perform long-term forecasting tasks. 
For example, there are more than 20,000 base stations in a medium-sized city, and each of them requires long-term time series forecasting, which will result in a single GPU being unable to handle such large-scale data. 

To address the above problems, fine-grained Long-term and Large-scale Wireless Traffic Forecasting (LL-WTF) methods at the node level urgently need to be developed.
In this paper, we delve into LL-WTF tasks, aiming to handle vast quantities of irregular erratic data and tackle the non-stability challenge inherent in LL-WTF tasks.
For large-scale WT datasets, we developed a Random Subgraph Sampling (RSS) algorithm, which, according to our findings, effectively addresses the training challenges of deep neural networks on large-scale irregular graph data.
The proposed RSS algorithm enables the disassembly of large-scale WT data into multiple small-scale subseries through random subgraphs sampling, allowing each to be processed on existing GPU devices.

For long-term WT forecasting, we utilize decomposition to deconstruct a time series into several more predictable components and leverage deep models to learn the time-varying patterns of each component separately.
Instead of treating decomposition as a preprocessing step \cite{taylor2018forecasting, oreshkin2019n, sen2019think} or embedding it in a learning module \cite{wu2021autoformer, zhou2022fedformer, liu2022non}, we use decomposition for supervised signals.
Concretely, we propose a Progressive Supervision method based on Label Decomposition (PSLD) to progressively learn the decomposed supervision signal to better cope with the non-stationary problem. PSLD utilizes label decomposition to obtain multiple easy-to-learn components, which are learned progressively at shallow layers and combined at deep layers.
Through the decomposition of label signals, PSLD can more easily learn various statistical properties and aperiodic trends of non-stationary time series.
It is worth noting that PSLD differs from the concept of PNNs \cite{rusu2016progressive}, which focus on incremental learning through lateral connections to previously learned networks, primarily aimed at addressing the problem of catastrophic forgetting. In contrast, our approach does not involve adding new network modules or preserving previous networks; instead, it emphasizes evolving the supervision signal itself as training progresses.

In summary, various decompositions and learners can be applied to PSLD, which can be regarded as a new paradigm that makes full use of decomposition to mine the time-varying properties of supervision signals and combines the powerful fitting capabilities of deep learning.
We implemented two versions of PSLD on MLP using classical statistical (Mean and Variance Decomposition, MVD) and STL \cite{cleveland1990stl}. 
Extensive experimental results show PSLD outperforms existing methods by a large margin, average 2\%, 4\%, and 11\% performance improvement on three LWTF datasets, respectively. 
As illustrated in Fig. \ref{fig1}, PSLD not only achieves state-of-the-art (SOTA) performance in terms of predictive accuracy but also maintains a leading position in processing speed.
In addition, we build an open source library for wireless traffic forecasting (WTFlib) to facilitate related research, which contains numerous SOTA methods and provides a strong benchmark.

The main contributions are summarized as follows.
\begin{itemize} 
  \item The long-term and large-scale wireless traffic forecasting (LL-WTF) task is introduced, and an open source library for WT forecasting (WTFlib) is built to facilitate related research.
  \item A Random Subgraph Sampling (RSS) algorithm is developed to disassemble large-scale WT data into multiple small-scale subseries, so that each of them can be processed on existing GPUs.
  \item A Label-Decomposition-based Progressive Supervision method (PSLD) is proposed, which utilizes label decomposition to obtain multiple easy-to-learn components. These components are learned progressively at shallow layers and combined at deep layers to address non-stationarity problem in LL-WTF.
  \item Extensive experiments over three large-scale WT datasets are conducted to verify the performance of the proposed methods. It is shown that PSLD improve the average performance of state-of-the-art (SOTA) methods by around 2\%, 4\%, and 11\%, respectively. 
  
\end{itemize}

The remainder of the paper is organized as follows:
In Section \ref{sec_rss}, we first introduce the RSS method.
Then, we introduce the architecture of PSLD in Section \ref{sec_PSLD}. 
Section \ref{sec_exp} presents the experimental results of the proposed method. 
Related works are presented in Section \ref{sec_rework}.
Finally, we conclude this paper in Section \ref{sec_con}.

\textbf{Definition}:
The purpose of LL-WTF is to use the observed value of $L_{in}$ historical moments to predict the missing value of $L_{out}$ future moments, which can be denoted as $Input \text{-}L_{in}\text{-}predict\text{-}L_{out}$. If the feature dimension of the series is denoted as $D$, its input data at time $t$ can be denoted as $X^t = \{s_1^t, \cdots, s_{L_{in}}^t | s_k^t \in \mathbb{R}^D  \}$, and its output at time $t$ can be denoted as $Y^t = \{s_{L_{in}+1}^t, \cdots, s_{L_{in}+L_{out}}^t | s_{L_{in}+k}^t \in \mathbb{R}^D  \}$. 
Then, we can predict $Y^t \in \mathbb{R}^{L_{out}\times D}$ by designing a model $\mathcal{F}$ given an input $X^t \in \mathbb{R}^{L_{in}\times D}$, which can be expressed as: $Y^t = \mathcal{F}(X^t)$. 
For denotation simplicity, the superscript $t$ will be omitted if it does not cause ambiguity in the context.

\begin{figure*}
  \centerline{\includegraphics[width=\textwidth]{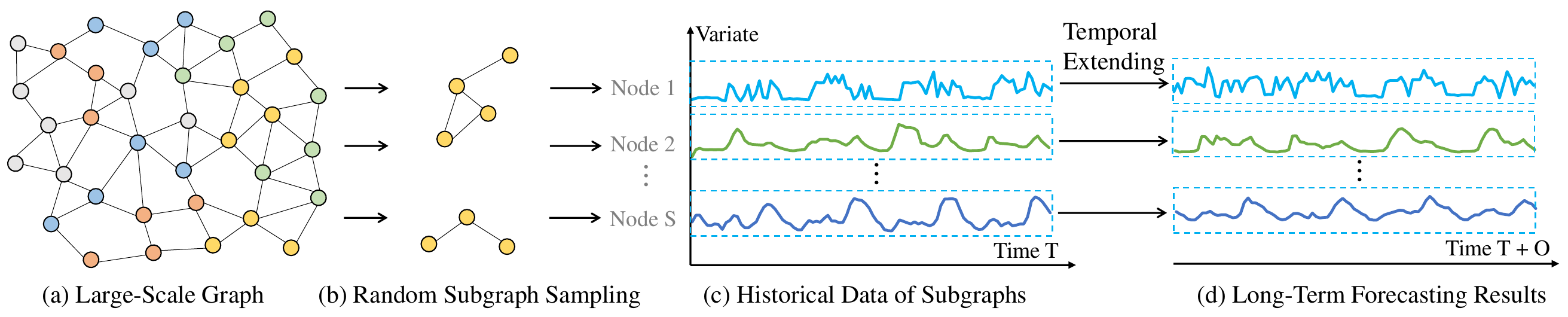}}
  \caption{Random Subgraph Sampling (RSS): For large-scale graph-structured data (a), a subgraph is randomly selected at each iteration (b). Through multiple sampling iterations, comprehensive coverage of the large-scale graph is achieved, ensuring the full utilization of its node information and structural data. During network training (repeated) and testing (non-repeated), long-term wireless traffic prediction (d) is implemented by temporally extending the historical data of the subgraph (c).}
  \label{fig_scene}
\end{figure*}

\section{Random Subgraph Sampling (RSS)}
\label{sec_rss}

Random Subgraph Sampling (RSS) aims to reduce the computational burden of training deep models on large graphs while maintaining the model's generalization performance. RSS randomly selects a subgraph $G_{sub}$ from the original large-scale graph $G=(V,E)$ to train the model $\mathcal{F}$, where $V$ indicates nodes and $E$ indicates edges.
As shown in Fig. \ref{fig_scene}.a, the original large-scale graph is disassembled into multiple small-scale subgraphs $\{X_{sub}, Y_{sub}, E_{sub}\}$ through RSS, so that each of them can be processed on existing GPU devices.
The following theorem proves that RSS offers comprehensive coverage of the original large-scale graph, which is an unbiased estimator of the true aggregated feature using the entire graph.
\begin{theorem}
  Subgraphs sampled by RSS are an unbiased estimator that leverages the true aggregate features of the entire graph.
  \label{th1}
\end{theorem}
The proof of Theorem \ref{th1} is given in Appendix \ref{sec_rss}.
Theorem \ref{th1} means that we can train a well-performing learning model by randomly sampling subgraphs of a large graph.
In Algorithm~\ref{alg:rss}, we provide a Python pseudocode implementation of RSS. 
We can perform random sampling of large-scale graphs in multiple batches through Algorithm 1.
\begin{algorithm}[t]
  \caption{Python Pseudocode Implementation of RSS}
  \label{alg:rss}
  \begin{algorithmic}[1]
    \REQUIRE graph $G=(V,E)$, the number of nodes $N_{node}$, the number of subgraphs $N_{sub}$, the length of the data $L_{data}$, input lengh $L_{in}$, output lengh $L_{out}$.
    \STATE  $N_{sub} \gets N_{node}/N_{sub}$
    \IF{Training}
      \STATE  $I_{index} \gets  $  random.shuffle([$1,...,N_{node}$])
    \ELSE
      \STATE  $I_{index} \gets $ [$1,...,N_{node}$]
    \ENDIF
    \STATE  $\hat{E}_{sub} \gets E[I_{index}]$
    \STATE  $E_{sub} \gets \hat{E}_{sub}[:,I_{index}]$
    \STATE  $L_{time} \gets L_{data} - L_{in} - L_{out} + 1$
	  \STATE  $V_{sub} \gets V[:, I_{index}]$
	  \STATE  $X_{sub} \gets$ [$V_{sub}$[$t$:$t$+$L_{in}$] for $t$ in range($L_{time}$)]
	  \STATE  $Y_{sub} \gets$ [$V_{sub}$[$t$+$L_{in}$:$t$+$L_{in}$+$L_{out}$] for $t$ in range($L_{time}$)]

    \RETURN random\_batch($X_{sub},Y_{sub},E_{sub}$)
  \end{algorithmic}
\end{algorithm}

\section{PSLD}
\label{sec_PSLD}

Decomposition is a powerful technique for analyzing and solving non-stationary problems in time series data \cite{cleveland1990stl, de2011forecasting, cheng2015time}. Decomposition addresses non-stationarity by breaking the time series into distinct components that can be more easily analyzed and modeled. The primary components typically considered are mean, variance, trend, seasonal, and residual (or noise) components.


\begin{figure*}
  \centerline{\includegraphics[width=\textwidth]{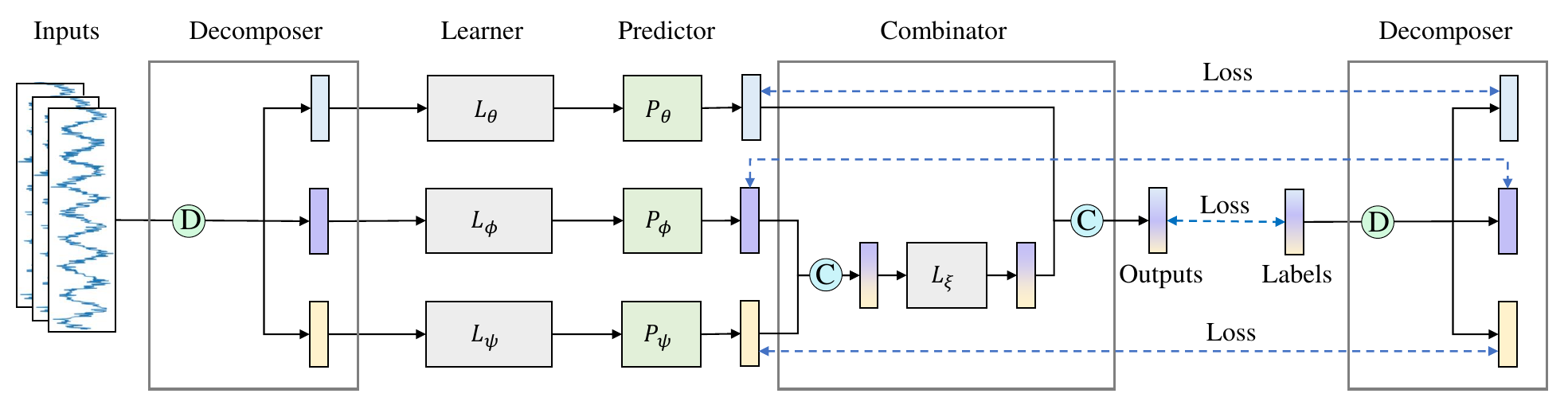}}
  \caption{The architecture of PSLD, which including three main parts: decomposer $D$, learner $L$ and predictor $P$, as well as combinator $C$. The decomposer is tasked with decomposing the input series into several easier to handle components. The learners are employed to model the nonlinear components of the prediction, while the predictor aligns the outputs of the individual learners with the decomposed components of the label. The combinator integrates the predictions of each component to derive the final output. During the learning process, each label component is back-propagated to the shallow layers to gradually supervise their learning process.}
  \label{fig_arch} 
\end{figure*}

In this section, we present PSLD to mitigate the non-stationary in WT data, which arise from trends, seasonal patterns, or other time-varying structures.
As shown in Fig. \ref{fig_arch}, the proposed PSLD consists of three main components: decomposer, learner and predictor, as well as combinator.

\subsection{Decomposer}

The decomposer is responsible for decomposing the target variable $Y$ into components that capture non-stationary and stationary behaviors respectively.
Many classic time series decomposition methods, including additive decomposition $Y = T + S + R$, multiplicative decomposition $Y = T\times S \times R$, and seasonal-trend decomposition $Y = T + S + R$ (e.g., STL), can be employed for label decomposition, 
where $T$ captures the long-term trend, $S$ captures the seasonal or cyclical patterns, and $R$ captures the residual stationary component.
In this paper, we utilize a mean-variance (hybrid additive and multiplicative) decomposition and an STL decomposition. 

\textbf{Mean-Variance Decomposer} (MVD): 
The mean-variance decomposer is a hybrid decomposition method that combines the advantages of both additive and multiplicative decompositions.
The decomposition of the target variable by MVD can be expressed as:
\begin{align}
  M & = \text{Mean}(Y, -1), \notag \\
  Y' & = Y - M, \notag \\
  V & = \text{Var}(Y', -1), \notag \\
  R & = Y' / V + \epsilon, \label{eq_mvd}
\end{align}
where $M$ and $V$ are the mean and variance of $Y$, respectively, $\epsilon$ is a small constant to avoid division by zero. 
The mean $M$, variance $V$, and variable $R$ in Eq. \ref{eq_mvd} are used to progressively supervise the learning process of the learner.

\textbf{STL Decomposer}: 
STL is a robust and flexible method that can decompose a complex time series into seasonal, trend, and residual components.
The decomposition of the target variable by the STL decomposer can be expressed as:
\begin{align}
  T & = \text{Move-Avg}(Y, \kappa_T), \notag \\
  Y' & = Y - T, \notag \\
  S & = \text{Move-Avg}(Y', \kappa_S), \notag \\
  R & = Y' - S, \label{eq_stl}
\end{align}
where Move-Avg is the moving average operation, $\kappa_T$ and $\kappa_S$ are the size of the trend and seasonal smoothing kernel, respectively.
Similar to MVD, the trend $T$, seasonal $S$, and residual $R$ components in Eq. \ref{eq_stl} are used to progressively supervise the learning process of the learner.

\subsection{Learner and Predictor}

To progressively guide the model's learning process, we employ different learners and predictors to give predictions for each component individually  at the model's shallow level.
To achieve this, we insert a decomposer at the input to decompose it into its component forms corresponding to those of the output, like
\begin{align}
  M_X & = \text{Mean}(X, -1), \notag \\
  X' & = X - M_X, \notag \\
  V_X & = \text{Var}(X', -1), \notag \\
  R_X & = X' / V_X + \epsilon, \label{eq_x_mvd}
\end{align}
Indeed, the learner and predictor can be any learnable models, where the learner extracts features and the predictor generates prediction results.
Let $L$ and $P$ be the learner and predictor parameterized by $\theta$, $\phi$ and $\psi$, respectively. 
For MVD, we have:
\begin{align}
  \hat{M} & = P_\theta(L_\theta(M_X)), \notag \\
  \hat{V} & = P_\phi(L_\phi(V_X)), \notag \\
  \hat{R} & = P_\psi(L_\psi(R_X)). \label{eq_learner}
\end{align}
The reason for separating the learner and the predictor is for greater flexibility. For example, the learner can be a deep neural network, while the predictor can use traditional regression methods. 
In some simple scenarios, employing a learner may not be necessary.

\subsection{Combinator}

The combiner integrates the predictions of each component to produce the final prediction result, which is the inverse of the decomposition process.
As shown in Fig. \ref{fig_arch}, 
The components are sequentially integrated in the combiner, which incorporates learnable parameters to coordinate the contributions of each component.
Specifically, for MVD, the prediction result of combiner can be expressed as:
\begin{align}
  \hat{Y} & = P_{\xi}\left(L_{\xi}(V \times R) + M \right) , \label{eq_cbn_mvd}
\end{align}
where $\xi$ is the parameter of the combiner. Similarly, For STL, the combiner can be expressed as:
\begin{align}
  \hat{Y} & = P_{\xi}\left(L_{\xi}(S + R) + T \right) . \label{eq_cbn_stl}
\end{align}

\subsection{Loss Function}

The loss function of PSLD comprises two parts: the prediction loss of each component generated by the predictor, and the final prediction loss produced by the combiner.
For MVD, the loss function of the components can be expressed as:
\begin{align}
  \mathcal{L}_{cpn} & = \mathcal{L}_{\theta}(\hat{M}, M) + \mathcal{L}_{\phi}(\hat{V}, V) + \mathcal{L}_{\psi}(\hat{R}, R) , \label{eq_loss_mvd}
\end{align}
and the loss function of the combiner can be expressed as:
\begin{align}
  \mathcal{L}_{cbn} & = \mathcal{L}_{\xi}(\hat{Y}, Y) . \label{eq_loss_cbn}
\end{align}
Thus,  the final loss function of PSLD can be expressed as:
\begin{align}
  \mathcal{L} & = \mathcal{L}_{cbn} + \lambda  \mathcal{L}_{cpn}, \label{eq_loss}
\end{align}
where $\lambda$ is a hyperparameter that controls the contribution of the component loss to the overall loss.
Likewise, the loss function of the STL decomposer is similar to that of MVD.

\subsection{Accelerating PSLD}

\begin{figure}
  \centerline{\includegraphics[width=\columnwidth]{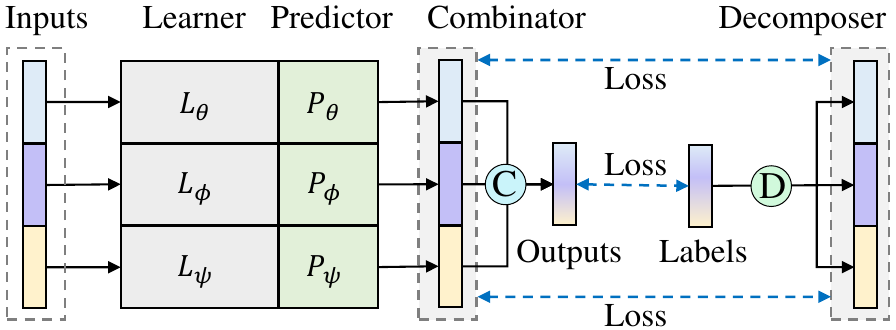}}
  \caption{The accelerated PSLD architecture. The decomposed components, along with the learners and predictors, are combined into longer vectors or wider models. This parallelization can greatly accelerate the training and inference process.
  }
  \label{fig_fast_arch} 
\end{figure}

It is observed that the learner and predictor in PSLD operate in parallel. If they share the same model structure, they can be merged to enhance computational efficiency.
As shown in Fig. \ref{fig_fast_arch}, the learners and predictors are merged into a single model, and their parameters are combined into a single set $[\theta, \phi, \psi ]$. 
This is equivalent to widening the original learner and predictor, which can be expressed as:
\begin{align}
  [\hat{M},\hat{V},\hat{R}] & = P_{[\theta, \phi, \psi ]}\left(L_{[\theta, \phi, \psi ]}([M_X, V_X, R_X]) \right) . \label{eq_fast_mvd}
\end{align}
Similarly, the component loss can be expressed as:
\begin{align}
  \mathcal{L}_{cpn} & = \mathcal{L}_{[\theta, \phi, \psi ]}([\hat{M},\hat{V},\hat{R}], [M, V, R]). \label{eq_fast_loss}
\end{align}
By employing Eq. \ref{eq_fast_loss}, we can train a single integrated model, which is typically more efficient than training multiple independent models due to the parallel operations.

\section{Experiments}
\label{sec_exp}

PSLD is extensively evaluated on the three widely used real-world WT datasets, including multiple mainstream WT applications such as C2TM, Milano and CBS. 


\subsection{Datasets}

1) C2TM \cite{Modeling15Chen} makes an analysis on week-long traffic generated by a large population of people in a median-size city of China.
C2TM analyses make use of request-response records extracted from traffic at the city scale, consisting of individuals' activities during a continuous week (actually eight days from Aug. 19 to Aug. 26, 2012), with accurate timestamp and location information indicated by connected cellular base stations. 
C2TM contains 13269 nodes, each with a length of 192. The ratio of training, validation and test datasets is 5:1:2.

2) Milano \cite{barlacchi2015multi} is provided by SKIL\footnote{http://jol.telecomitalia.com/jolskil} of Telecom Italia. The dataset was collected from November 1, 2013, to January 1, 2014, and the data is aggregated into 10-minute intervals over the whole city of Milan (62 days, 300 million records, about 19 GB).
Milano records the time of the interaction and the specific radio base station that managed it. 
The dataset contains 10,000 nodes, and the length of each node is 1498. The ratio of training, validation and test datasets is 10:2:3.

3) CBS is collected from all base station in a certain city in China, including up-link and
down-link communication trafﬁc data. The dataset is collected from June 8 to July 26, 2019, with a temporal interval of 60 minutes over the whole city (94 days, 51.9 million records, about 17.1 GB). 
CBS contains 4454 nodes, each with a length of 4032. The ratio of training, validation and test datasets is 3:1:1.

\subsection{Implementation Details}

For the WT datasets, we initially employ the RSS algorithm to perform data sampling. In this algorithm, we random decompose the entire large graph into 24 sub-graphs at each sampling, as specified in Algorithm 1.
For the learner in PSLD, we employ a 2-layer MLP with ReLU as the activation function. For the predictor, we utilize a linear layer. 
We selected this architecture due to its simplicity, computational efficiency, and ability to model nonlinear relationships effectively, which aligns well with the structure of our dataset.
While CNNs are particularly suited for capturing spatial dependencies and GNNs are ideal for graph-based data, our problem does not exhibit strong spatial structures that would benefit significantly from CNNs, nor does it require the additional complexity that GNNs bring. 
Therefore, we found the MLP to be the most appropriate choice for balancing performance and computational cost in this case.
We implemented two versions of PSLD: one utilizing the MVD (denoted as  PSLD-MVD) and the other employing the STL decomposer (denoted as PSLD-STL).

The data preprocessing of the training datasets $X_{train}$ is calculated as:
\begin{align}
    M = \frac{1}{N} \sum_{i=1}^{N} X_{train}[i] \notag \\
    S = \sqrt{\frac{1}{N} \sum_{i=1}^{N} (X_{train}[i] - M)^2} \notag \\
    \tilde{X}_{train} = (X_{train} - M) / S \notag \\
    \tilde{X}_{test} = (X_{test} - M) / S  
\end{align}
where $N$ is the number of samples, $M$ and $S$ are the mean and standard deviation of the training set, respectively. It is worth noting that the data preprocessing step is consistent across all datasets, and we did not use any other data preprocessing techniques.

Furthermore, the number of hidden units is set to 128. The dropout rate is set to 0.05. 
The model is trained using the ADAM \cite{kingma2014adam} optimizer and L2 loss.
The learning rate is set to 0.0001. 
The batch size is set to 1 for large-scale graph WT datasets, and the maximum number of epochs is set to 10, which is consistent with \cite{Zhou2021Informer, wu2021autoformer}
The model is trained on a single NVIDIA Tesla V100 GPU with 32GB memory.
And the software environment and dependencies include Python 3.10 and PyTorch 2.0.
To facilitate the replication experiment, we provide links to easily accessible preprocessed datasets on \href{https://drive.google.com/drive/folders/1Yu-_PXzOSEqfTZbAO8cKptVXU3Cz9o3T?usp=sharing}{Google Cloud}.

\subsection{Baselines}
For comparison purpose, 13 SOTA Transformer-based models are adopted as baselines, including FreTS \cite{yi2024frequency}, TimeMachine \cite{ahamed2024timemachine}, FourierGNN \cite{yi2024fouriergnn},  PatchTST \cite{nietime2023ICLR}, STID \cite{shao2022spatial}, Periodformer \cite{liang2023does}, DLinear \cite{zeng2023transformers}, FEDformer \cite{zhou2022fedformer}, Autoformer \cite{wu2021autoformer}, Informer \cite{Zhou2021Informer}, Transformer \cite{NIPS2017_Transformer}, Mvstgn \cite{yao2021mvstgn}, and GWNet \cite{wu2019graph}.
All models use the same settings as PSLD, for example, the RSS algorithm is used in the data sampling process.
The models used in the experiments are evaluated over a wide range of prediction lengths to compare performance on different future horizons: 4, 5, 6, 7, and 8 for the C2TM dataset (because the length of this dataset is short.), and 24, 36, 48, and 72 for others. The experimental settings are the same for all LL-WTF tasks.
Other graph neural networks, such as DyDgcrn \cite{li2023dynamic}, were not adopted since they could not implement the LL-WTF task due to their significant computational and storage overhead.
The experimental results are shown in Table \ref{tb2}.

\begin{table*}
    \centering
    \caption{LL-WTF results on six benchmark datasets. A lower MSE or MAE indicates a better performance, and the best results are highlighted in bold.}
    \label{tb2}
    \resizebox{\textwidth}{!}
  {
    \begin{tabular}{cc| ccccc| cccc| cccc}
      \toprule
    \multirow{2}{*}{Methods}      & Datasets & \multicolumn{5}{c|}{C2TM}                   & \multicolumn{4}{|c}{Milano}        & \multicolumn{4}{|c}{CBS}           \\ 
                                  & Lengths                        & 4      & 5      & 6      & 7      & 8      & 24     & 36     & 48     & 72     & 24     & 36     & 48     & 72     \\ \midrule
    \multirow{2}{*}{PSLD-MVD}  & MSE                     & \bf 9.057  & \bf 9.161  & \bf 9.298  & \bf 9.431  & \bf 9.383  &\bf  0.633  & \bf 0.775  & \bf 0.907  & \bf 1.184  & \bf 1.641  & 1.768  & 1.822  & \bf 1.955  \\ 
                                  & MAE                     & \bf 0.171  & 0.174  & \bf 0.171  & 0.317  & 0.169  & \bf 0.249  & \bf 0.269  & 0.285  & \bf 0.301  & \bf 0.625  & \bf 0.644  & \bf 0.646  & \bf 0.660  \\ \midrule
    \multirow{2}{*}{PSLD-STL} & MSE                     & 9.084  & 9.183  & 9.328  & 9.447  & 9.400  & 0.708  & 0.870  & 1.062  & 1.366  & 1.644  & \bf 1.759  & \bf 1.821  & 1.965  \\
                                  & MAE                     & 0.174  & \bf 0.172  & 0.172  & \bf 0.169  & \bf 0.167  & 0.275  & 0.294  & 0.315  & 0.343  & 0.633  & 0.649  & 0.653  & 0.669  \\ \midrule
    \multirow{2}{*}{FreTS}        & MSE                     & 9.058  & \bf 9.161  & 9.302  & 9.440  & 9.397  & 0.702  & 0.861  & 0.966  & 1.193  & 1.784  & 1.952  & 2.025  & 2.053  \\
                                  & MAE                     & 0.189  & 0.192  & 0.186  & 0.189  & 0.191  & 0.276  & 0.293  & 0.305  & 0.320  & 0.682  & 0.709  & 0.715  & 0.719  \\ \midrule
    \multirow{2}{*}{TimeMachine}  & MSE                     & 9.083  & 9.187  & 9.333  & 9.475  & 9.432  & 0.644  & 0.788  & 0.950  & 1.264  & 1.715  & 1.867  & 1.915  & 2.052  \\
                                  & MAE                     & 0.176  & 0.175  & 0.178  & 0.178  & 0.177  & 0.250  &\bf 0.269 &\bf 0.282& 0.303 & 0.663& 0.685  & 0.685  & 0.699  \\ \midrule
    \multirow{2}{*}{FourierGNN}   & MSE                     & 9.064  & 9.164  & 9.306  & 9.447  & 9.400  & 0.843  & 1.000  & 1.123  & 1.450  & 2.601  & 2.713  & 2.812  & 1.193  \\
                                  & MAE                     & 0.204  & 0.199  & 0.198  & 0.188  & 0.187  & 0.311  & 0.330  & 0.340  & 0.360  & 0.826  & 0.842  & 0.854  & 0.320  \\ \midrule
    \multirow{2}{*}{PatchTST}     & MSE                     & 9.226  & 9.340  & 9.509  & 9.504  & 9.561  & 0.662  & 0.832  & 0.929  & 1.222  & 1.870  & 2.036  & 2.113  & 2.084  \\
                                  & MAE                     & 0.195  & 0.197  & 0.201  & 0.182  & 0.192  & 0.254  & 0.273  & \bf 0.282  & 0.302  & 0.706  & 0.736  & 0.735  & 0.697  \\ \midrule
    \multirow{2}{*}{STID}         & MSE                     & 9.213  & 9.769  & 9.625  & 9.925  & 9.560  & 0.730  & 0.840  & 0.985  & 1.248  & 2.066  & 2.038  & 1.986  & 2.416  \\
                                  & MAE                     & 0.375  & 0.339  & 0.353  & 0.509  & 0.387  & 0.367  & 0.382  & 0.377  & 0.404  & 0.744  & 0.738  & 0.706  & 0.797  \\ \midrule
    \multirow{2}{*}{MVSTGN}       & MSE                     & 9.087  & 9.195  & 9.321  & 9.433  & 9.395  & 1.372  & 1.443  & 2.656  & 2.928  & 3.001  & 2.442  & 3.013  & 3.442  \\
                                  & MAE                     & 0.352  & 0.360  & 0.357  & 0.329  & 0.337  & 0.524  & 0.490  & 0.559  & 0.731  & 0.813  & 0.783  & 0.861  & 0.907  \\ \midrule
    \multirow{2}{*}{GWNet}        & MSE                     & 9.178  & 9.288  & 9.431  & 9.443  & 9.420  & 1.210  & 1.603  & 2.151  & 3.413  & 1.893  & 2.060  & 2.074  & 2.225  \\
                                  & MAE                     & 0.610  & 0.617  & 0.624  & 0.409  & 0.412  & 0.480  & 0.526  & 0.564  & 0.573  & 0.714  & 0.760  & 0.739  & 0.754  \\ \midrule
    \multirow{2}{*}{Periodformer} & MSE                     & 11.584 & 11.663 & 11.822 & 11.883 & 11.818 & 1.341  & 1.457  & 1.583  & 1.892  & 5.139  & 5.491  & 5.335  & 5.515  \\
                                  & MAE                     & 1.001  & 0.980  & 0.975  & 0.938  & 0.923  & 0.571  & 0.539  & 0.542  & 0.540  & 1.213  & 1.251  & 1.231  & 1.235  \\ \midrule
    \multirow{2}{*}{DLinear}      & MSE                     & 12.702 & 12.580 & 12.990 & 13.457 & 12.945 & 0.898  & 1.031  & 1.106  & 1.338  & 2.414  & 2.576  & 2.584  & 2.700  \\
                                  & MAE                     & 0.516  & 0.464  & 0.494  & 0.456  & 0.556  & 0.359  & 0.370  & 0.392  & 0.403  & 0.781  & 0.805  & 0.806  & 0.815  \\ \midrule
    \multirow{2}{*}{FEDformer}    & MSE                     & 11.587 & 11.631 & 11.679 & 11.763 & 11.544 & 1.836  & 1.836  & 1.961  & 2.274  & 6.048  & 6.368  & 6.474  & 6.434  \\
                                  & MAE                     & 0.969  & 0.943  & 0.914  & 0.881  & 0.825  & 0.734  & 0.734  & 0.762  & 0.772  & 1.424  & 1.469  & 1.472  & 1.474  \\ \midrule
    \multirow{2}{*}{Autoformer}   & MSE                     & 11.680 & 11.721 & 11.828 & 11.940 & 11.832 & 1.796  & 2.652  & 2.571  & 2.617  & 6.107  & 6.345  & 6.089  & 6.078  \\
                                  & MAE                     & 1.030  & 1.010  & 0.994  & 0.981  & 0.951  & 0.733  & 0.999  & 0.955  & 0.898  & 1.453  & 1.472  & 1.432  & 1.412  \\ \midrule
    \multirow{2}{*}{Informer}     & MSE                     & 9.401  & 9.513  & 9.634  & 9.894  & 9.724  & 20.268 & 20.236 & 20.292 & 20.394 & 15.156 & 15.125 & 15.196 & 15.219 \\
                                  & MAE                     & 0.581  & 0.581  & 0.582  & 0.670  & 0.577  & 2.137  & 2.111  & 2.098  & 2.079  & 2.365  & 2.362  & 2.370  & 2.372  \\ \midrule
    \multirow{2}{*}{Transformer}  & MSE                     & 9.373  & 9.474  & 9.612  & 9.796  & 9.712  & 20.049 & 20.019 & 20.113 & 20.243 & 14.622 & 15.097 & 15.137 & 15.156 \\
                                  & MAE                     & 0.584  & 0.583  & 0.582  & 0.639  & 0.580  & 1.988  & 1.948  & 1.941  & 1.906  & 2.339  & 2.351  & 2.352  & 2.392  \\ \bottomrule

    \end{tabular}
  }
    \end{table*}

\subsection{Main Results}

The results for LL-WTF are summarized in Table \ref{tb2}, PSLD achieves the consistent SOTA performance in all datasets and prediction length settings. Compared to PatchTST and STID, the proposed PSLD yields an overall {\bf 5.4\%} and {\bf 8.7\%} relative MSE reduction, respectively. 
Specifically, for the input-8-predict-5 setting in C2TM, PSLD-MVD gives {\bf 1.9\%} (9.340$\rightarrow$9.161, compared to PatchTST) and {\bf 6.2\%} (9.769$\rightarrow$9.161, compared to STID) MSE reduction.
For the input-8-predict-6 setting in C2TM, PSLD-MVD gives {\bf 2.2\%} (9.509$\rightarrow$9.298, compared to PatchTST) and {\bf 3.4\%} (9.625$\rightarrow$9.298, compared to STID) MSE reduction.

For input-36-predict-36 setting, PSLD-MVD has {\bf 6.9\%} (0.832$\rightarrow$0.775) and {\bf 7.7\%} (0.840$\rightarrow$0.775) MSE reduction in Milano, 
{\bf 13.0\%} (2.036$\rightarrow$1.772) and {\bf 13.1\%} (2.038$\rightarrow$1.772) in CBS, etc. 
PSLD also outperforms other state-of-the-art models by a large margin on other settings, such as the latest Long-Term Series Forecasting (LTSF) models, including including FreTS \cite{yi2024frequency}, TimeMachine \cite{ahamed2024timemachine}, and FourierGNN \cite{yi2024fouriergnn}.
All the above experimental results have verified that the proposed PSLD can achieve better prediction performance on different datasets with varying horizons, implying its superiority on LL-WTF tasks. 
The ability to decompose the time series and model each component separately makes our proposed method robust to non-stationary data. This advantage is reflected in the lower error metrics compared to models that do not explicitly handle non-stationarity.

\subsection{Comparative Analysis} 

Notably, several pioneering models have also achieved competitive performance on certain datasets under particular settings. 
For instance, Informer demonstrates relatively poor performance on the WT datasets with different horizons settings.
This is due to the substantial presence of non-stationarity in each variate of the WT datasets.
This renders the KL-divergence-based ProbSparse Attention, as adopted in Informer, ineffective on this non-stationarity dataset.
Additionally, linear-based methods (e.g., STID and DLinear) have demonstrated promising results on the WT datasets with various horizons setting, while GNN-based methods (e.g., MVSTGN and GWNet) have yielded favorable results on these WT datasets. 
This phenomenon can be ascribed to a twofold interplay of factors. Previously, WT data exhibits stronger non-stationarity, including increased non-periodicity and noise, which directly affects the model's generalization.
Secondarily, other models exhibit a propensity to overfit non-stationary WT characterized by aperiodic fluctuations.
Remarkably, PSLD adeptly mitigates non-stationary challenges in LL-WTF forecasting, thereby enhancing its overall performance.
Particularly on the WT datasets with large-scale variables, PSLD achieves superior performance by adopting label decomposition to obtain multiple easy-to-learn components. These components are learned progressively at shallow layers and combined at deep layers to effectively address the non-stationary problem posed by LL-WTF tasks.

\subsection{Ablation Study}

\begin{figure*}
  \centerline{\includegraphics[width=\textwidth]{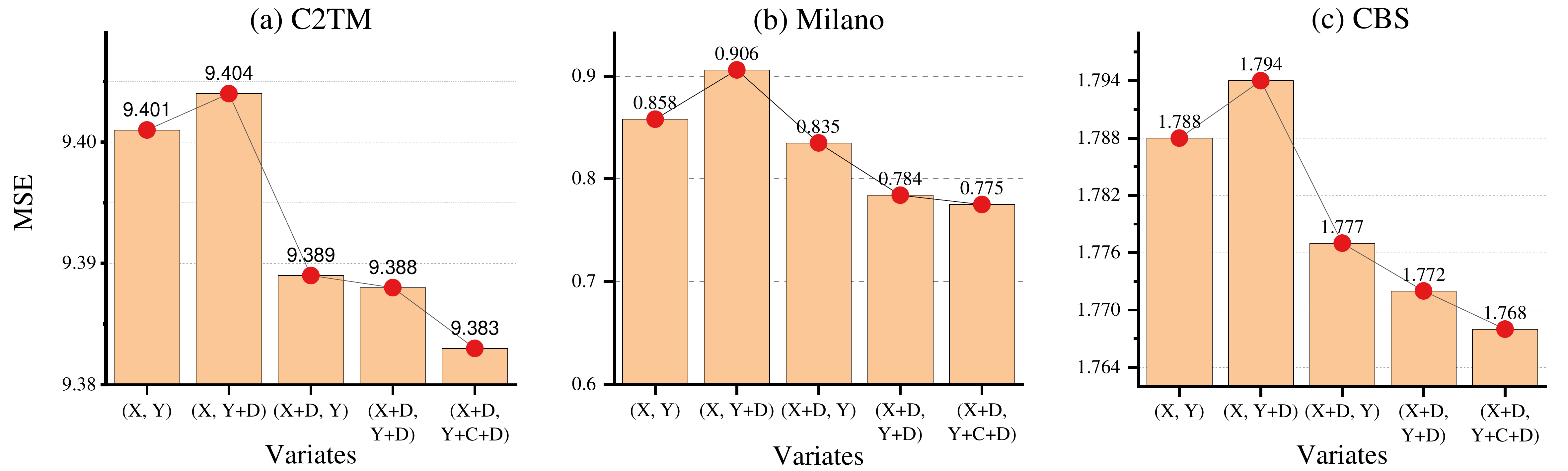}}
  \caption{Ablation studies on various components of PSLD. All results are averaged across all prediction lengths. The variables X and Y represent the input and output streams, while the signs `D' and `C' denote the decomposer and combinator when they are adopted for input, label or output processing.}
  \label{fig_variates} 
\end{figure*}

\begin{figure*}
  \centerline{\includegraphics[width=\textwidth]{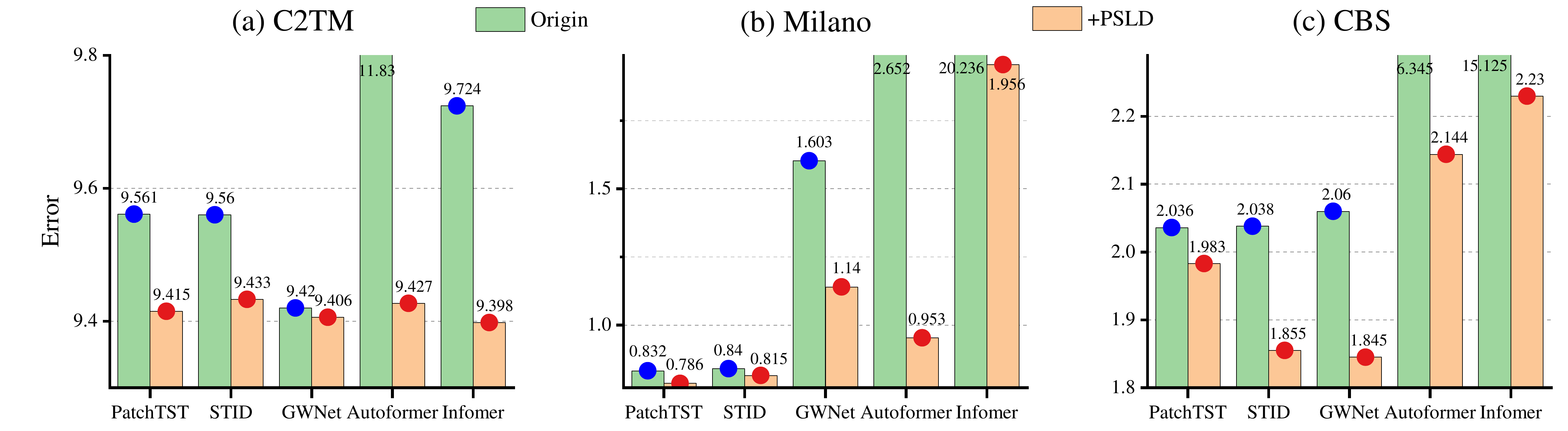}}
  \caption{Ablation studies of PSLD with other SOTA models. The tick labels
  of the X-axis are the different forecasting models. It is evident that the model's error is significantly reduced after incorporating PSLD.}
  \label{fig_models} 
\end{figure*}

\textbf{Component Analysis}:
To validate the effectiveness of PSLD, we conduct comprehensive ablation studies encompassing both component replacement and component removal experiments, as shown in Fig. \ref{fig_variates}.
We utilize letters `D' and `C' to denote the utilization of decomposer or combinator during the aggregation process of the input $X$ or output $Y$ streams. 
In cases involving only input streams, it becomes evident that the model's average performance is superior when employing decomposer on $X$ (X+D, Y) compared to when employing it on $Y$ (X, Y+D).
E.g., on the Milano dataset, forecast error is reduced by {\bf 7.8\%} (0.906$\rightarrow$0.835).
Moreover, the introduction of decomposers to the input and output streams respectively, further enhances the model's performance, e.g., forecast error on the CBS dataset is reduced by {\bf 0.9\%} (1.788$\rightarrow$1.772) as shown in Fig. \ref{fig_variates}(c).
Afterward, incorporating combinator (C) into the model holds the potential to improve predictive performance again, e.g., on the C2TM dataset, forecast error is reduced by {\bf 0.8\%} (0.784$\rightarrow$0.775).
In summary, integrating the decomposer and combinator components has the potential to significantly boost the model's performance across the board.

\textbf{Subgraph Partitioning}:
We conducted ablation experiments with different subgraph sizes on two large-scale graph datasets: Milano and CBS.
For these datasets, we initially employ the Random Subgraph Sampling (RSS) algorithm to perform data sampling. In RSS algorithm, we random decompose the entire large graph into 12, 24 and 32 subgraphs at each sampling. 
As shown in Table \ref{r1_q3:comp_subgraph}, the experimental findings indicate that the model's performance is stable across different subgraph sizes, reinforcing the robustness of the subgraph partitioning.

\begin{table}
  \centering
  \caption{Ablation study of subgraph partitioning.}
  \label{r1_q3:comp_subgraph}
  \resizebox{\columnwidth}{!}{
  \begin{tabular}{c|ccc|ccc}
    \toprule
           & \multicolumn{3}{c|}{Milano} & \multicolumn{3}{c}{CBS} \\ \hline
  Models   & 12      & 24      & 32     & 12     & 24     & 32    \\  \hline
  PSLD     & 0.633   & 0.633   & 0.632  & 1.642  & 1.641  & 1.641 \\
  PatchTST & 1.164   & 1.162   & 1.161  & 1.871  & 1.870  & 1.869 \\
  STID     & 1.227   & 1.230   & 1.228  & 2.065  & 2.066  & 2.067 \\
  DLinear  & 1.601   & 1.598   & 1.597  & 2.413  & 2.414  & 2.414 \\  \bottomrule
  \end{tabular}}
\end{table}

\subsection{Versatility}

To examine the versatility of PSLD as a comprehensive framework, we integrate the PSLD mechanism into various models to observe enhancements in their performance.
As shown in Fig. \ref{fig_models}, after harnessing the newly invented PSLD within other models, their performance exhibited considerable improvement. 
For example, the average MSE of PatchTST on the C2TM and Milano datasets witnessed a reduction of {\bf 1.5\%} (9.561$\rightarrow$9.415), {\bf 5.5\%} (0.832$\rightarrow$0.786), and {\bf 2.6\%} (2.036$\rightarrow$1.983), respectively, surpassing the original model by a large margin.
Furthermore, both STID, GWNet and other Attention-based models demonstrate commendable performance on the aforementioned datasets.
The conducted experiments suggest that PSLD can serve as a versatile architecture, amenable to the integration of novel modules, thereby facilitating the enhancement of performance in the domain of TS forecasting.

\subsection{Generalizability}

To verify the generalization of PSLD, we include comparative experiments with other Long-Term Sequence Forecasting (LTSF) models across multiple LTSF benchmark datasets, including ETT \cite{Zhou2021Informer}, ILI \footnote[1]{https://gis.cdc.gov/grasp/fluview/fluportaldashboard.html}, Weather \footnote[2]{https://www.bgc-jena.mpg.de/wetter}, Electricity \footnote[3]{https://archive.ics.uci.edu/ml/datasets/Electricity}, and Traffic \footnote[4]{http://pems.dot.ca.gov}.
Comprehensive forecasting results are presented in Table \ref{r2_q2:comp_other_datasets}, with the best results highlighted in bold. The lower MSE indicates the more accurate prediction result.
The proposed PSLD demonstrates superior performance across various LTSF benchmark datasets. 
Notably, PatchTST, previously the best model for weather datasets, fails in many cases on LTSF datasets. This can be attributed to the highly volatile nature of the data, which may cause the patching mechanism of PatchTST to lose focus on specific localities necessary for managing rapid fluctuations.
By contrast, PSLD utilizes label decomposition to obtain multiple easy-to-learn components, which are learned progressively at shallow layers and combined at deep layers. 
Therefore, the proposed method can progressively learn the decomposed supervision signal, allowing it to better address the non-stationary problem of LTSF.

\begin{table}
    \centering
    \caption{Comparisons of various LTSF models across multiple LTSF benchmark datasets, evaluated based on MSE under input-96-output-96 setting (input-36-output-24 for ILI).}
    \label{r2_q2:comp_other_datasets}
    \resizebox{\columnwidth}{!}{
    \begin{tabular}{cccccc}
      \toprule
                   & ETTm2 & ILI   & Weather & Electricity & Traffic \\ \hline
        PSLD       & \bf 0.164 & \bf 2.872 & 0.177   & \bf 0.180       & \bf 0.400     \\
        PatchTST   & 0.166 & 2.896 &\bf 0.152   & 0.195       & 0.544   \\
        DLinear    & 0.167 & 2.915 & 0.196   & 0.197       & 0.65    \\
        FEDformer  & 0.204 & 3.228 & 0.217   & 0.193       & 0.587   \\
        Autoformer & 0.255 & 3.228 & 0.266   & 0.201       & 0.613   \\ \bottomrule
    \end{tabular}
  }
  \end{table}

\subsection{Efficiency}

As shown in Table \ref{r1_q2:comp_times}, the proposed PSLD demonstrates superior performance, providing significantly faster inference compared to most models. Although the purely linear DLinear has the fastest training and inference speed, it sacrifices accuracy.
Mvstgn, STID and other Transformer-based models have the highest complexity, significantly slower training and inference times, and poorer accuracy.
In summary, PSLD demonstrates a balance of speed (training and inference), accuracy (lowest MSE), and moderate computational efficiency. Therefore, PSLD is optimal for scenarios requiring both high accuracy and computational efficiency.

\begin{table}[!ht]
    \centering
    \caption{Comparison of model's training time (Seconds/Epoch), inference time (Seconds/Epoch), parameters, FLOPs and MSE.}
    \label{r1_q2:comp_times}
    \resizebox{\columnwidth}{!}{
    \begin{tabular}{cccccc}
      \toprule
        Models       & Training (S) & Inference (S) & Params (M) & Flops (G) & MSE    \\ \hline
        Transformer  & 28.46             & 8.86               & 12.0       & 2.3       & 20.549 \\
        Informer     & 36.75             & 12.81              & 12.79      & 2.13      & 20.768 \\
        Autoformer   & 37.6              & 15.18              & 12.63      & 2.43      & 2.296  \\
        FEDformer    & 80.89             & 31.28              & 12.63      & 2.43      & 2.336  \\
        PatchTST     & 249.08            & 61.72              & 6.54       & 249.49   & 1.162  \\
        Periodformer & 26.55             & 13.79              & 8.95       & 1.73      & 1.841  \\
        Mvstgn       & 182.78            & 67.14              & 5.05       & 13.84     & 1.872  \\
        GWNet        & 38.74             & 15.19              & 3.25       & 0.524     & 1.710   \\
        STID         & 18.64             & 7.98               & 8.53       & 17.721    & 1.230   \\
        DLinear      & 2.51              & 0.95               & 1.8        & 0.0367    & 1.598  \\
        PSLD         & 6.86              & 2.38               & 2.0        & 0.232     & 0.633  \\ \bottomrule
    \end{tabular}
  }
  \end{table}

\subsection{Visualization the Learning Process of MVD}

\begin{figure*}
  \centerline{\includegraphics[width=\textwidth]{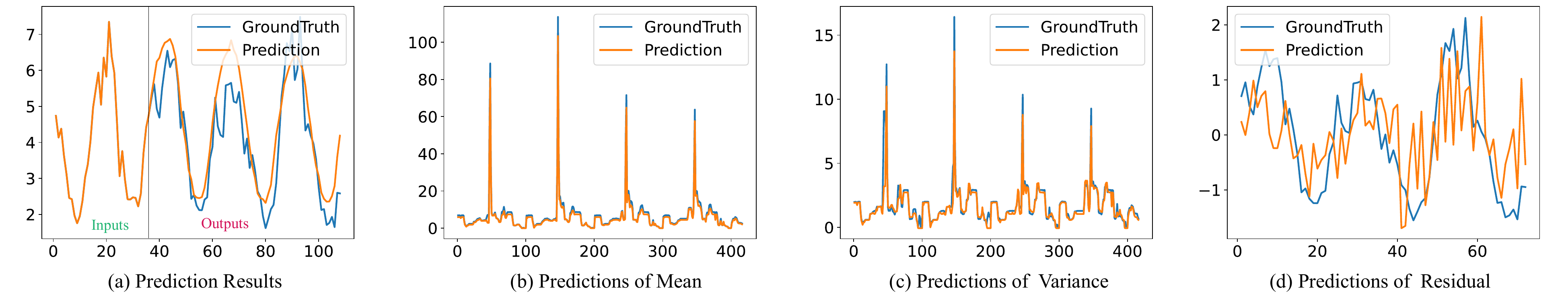}}
  \caption{Visualization of PSLD prediction results utilizing MVD. Prediction cases from the Milano dataset under the input-36-predict-72 setting.}
  \label{fig_vis_mvd} 
\end{figure*}

\begin{figure*}
  \centerline{\includegraphics[width=\textwidth]{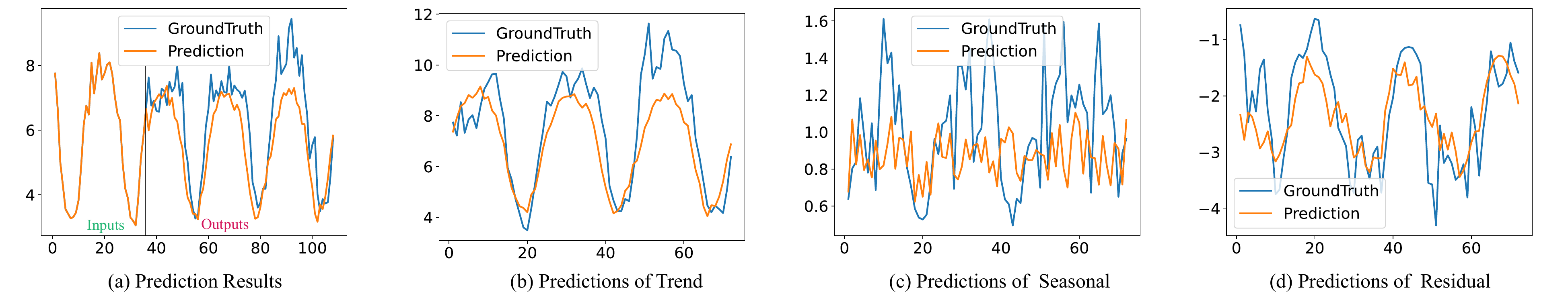}}
  \caption{Visualization of PSLD prediction results utilizing STL. Prediction cases from the Milano dataset under the input-36-predict-72 setting.}
  \label{fig_vis_stl} 
\end{figure*}

The intrinsic characteristic of PSLD framework lies in the alignment of the output from each component with the shape of the label's decompositions. This alignment, in turn, expedites the decomposition and visualization of the model's learning process.
As depicted in Figs. \ref{fig_vis_mvd} and \ref{fig_vis_stl}, the output of each component in PSLD is visualized. Fig. \ref{fig_vis_mvd}(a) shows the prediction results of MVD, Fig. \ref{fig_vis_mvd}(b) displays the predictions of the mean, Fig. \ref{fig_vis_mvd}(c) illustrates the predictions of the variance, and Fig. \ref{fig_vis_mvd}(d) presents the predictions of the residual.
It becomes evident that each component discerns and assimilates meaningful patterns within the series: Decomposing $Y$ into mean and variance provides clear insights into the underlying structure of the data. The mean captures the central tendency, while the variance captures the spread or dispersion.
For instance, in Fig. \ref{fig_vis_mvd}, despite the differences in the amplitudes of the series, the mean and variance components are accurately fitted to their groudn-truth values.
Specifically, comparing Figs. \ref{fig_vis_mvd}(a) and \ref{fig_vis_mvd}(b), for non-stationary series with varying mean and variance, each block must learn salient patterns: The mean can capture trends and seasonality, while the variance can account for changes in variability over time.
In addition, by predicting the variance, it is possible to quantify the uncertainty of predictions, leading to more robust and reliable forecasting. We leave this for future work.

\subsection{Visualization the Learning Process of STL}

The prediction results of STL along with the trend-season and residual components is visualized in Fig. \ref{fig_vis_stl}.
However, owing to the employment of different decomposers, the learning patterns of the components in Fig. \ref{fig_vis_stl} exhibit slight variations compared to those in Fig. \ref{fig_vis_mvd}.
STL decomposes the time series into trend, seasonal, and residual components. By isolating the trend component, the model can focus on capturing long-term patterns separately from seasonal fluctuations and random noise, which is particularly useful in non-stationary time series.
Meanwhile, STL provides a clear separation of trend, seasonal, and irregular components, making it easier to interpret the underlying structure of the time series. This separation helps stakeholders understand the contributions of different factors to the overall behavior of the series.
By understanding the trend and seasonal components separately, businesses can derive actionable insights, such as identifying growth trends or understanding the impact of seasonal demand fluctuations.

\subsection{Visualization of LL-WTF Results}

\begin{figure*}
  \centerline{\includegraphics[width=\textwidth]{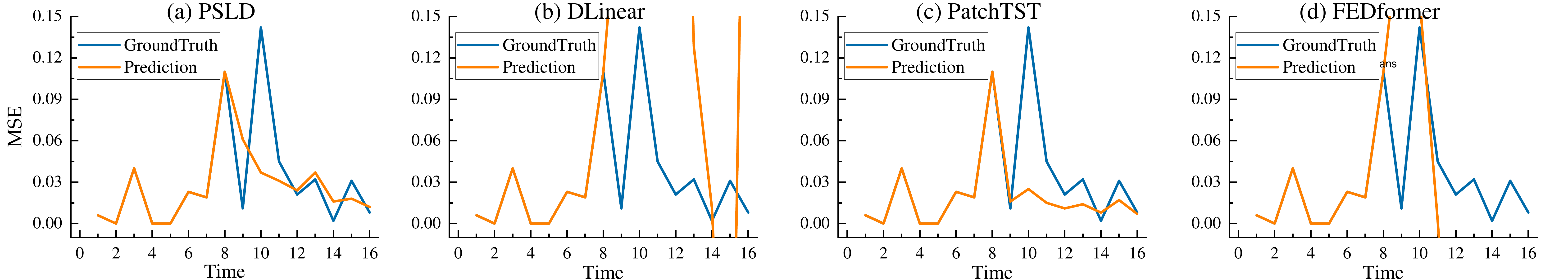}}
  \caption{Visualization of the prediction results across multiple models. Prediction cases from the C2TM dataset under the input-8-predict-8 setting.}
  \label{fig_vis_c2tm} 
\end{figure*}

\begin{figure*}
  \centerline{\includegraphics[width=\textwidth]{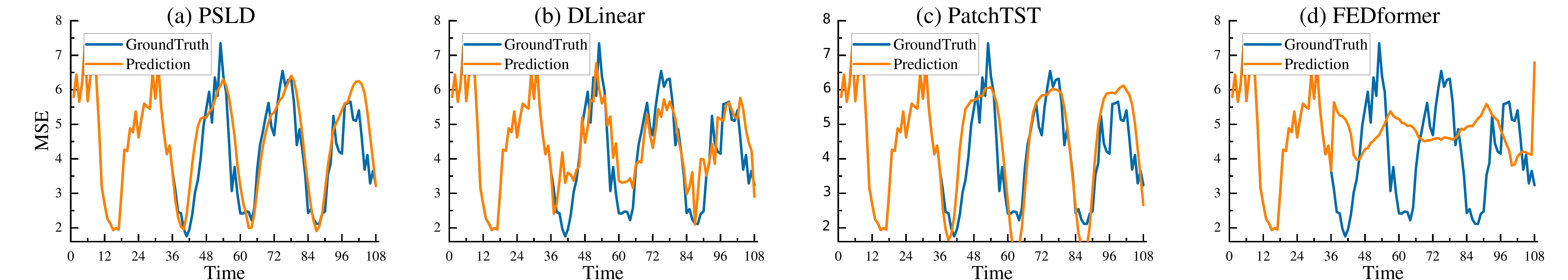}}
  \caption{Visualization of the prediction results across multiple models. Prediction cases from the Milano dataset under the input-36-predict-36 setting.}
  \label{fig_vis_internet} 
\end{figure*}

For clarity and comparison among different models, we present supplementary prediction showcases for three representative datasets in Figs. \ref{fig_vis_c2tm} and \ref{fig_vis_internet}. Visualization of different models for qualitative comparisons. Prediction cases from
the C2TM and Milano datasets. Two showcases correspond to predictions made by the following models: DLinear \cite{zeng2023transformers}, PatchTST \cite{nietime2023ICLR}, and FEDformer \cite{zhou2022fedformer}. Among the various models considered, the proposed PSLD stands out for its ability to predict future series variations with exceptional precision, demonstrating superior performance.

\section{Related Work}
\label{sec_rework}

\subsection{Classical Models for WT Forecasting}

Early classical methods \cite{piccolo1990distance,gardner1985exponential, li2010parsimonious} are widely applied to WT forecasting because of their well-defined theoretical guarantee and interpretability. For example, ARIMA \cite{piccolo1990distance} initially transforms a non-stationary time series into a stationary one via difference, and subsequently approximates it using a linear model with several parameters. 
Exponential smoothing \cite{gardner1985exponential} predicts outcomes at future horizons by computing a weighted average across historical data. In addition, some regression-based methods, e.g., random forest regression \cite{liaw2002classification} and support vector regression \cite{castro2009online}, etc., are also applied to WT forecasting. 
These methods are straightforward and have fewer parameters to tune, making them a reliable workhorse for WT forecasting. 
However, their shortcoming is insufficient data fitting ability, especially for high-dimensional series, resulting in limited performance.

\subsection{Deep Models for WT Forecasting}

The advancement of deep learning has greatly boosted the progress of WT forecasting. 
Specifically, 
the authors in \cite{huang2017study} treat the wireless traffic dataset as gridded data and utilize CNNs \cite{lecun1998gradient} and RNNs \cite{connor1994recurrent} to capture spatial and temporal correlations, respectively.
The authors in \cite{STDenseNet2018} investigate the spatial and temporal dependence of wireless traffic among different cells, where the spatiotemporal DenseNet \cite{huang2017densely} based prediction framework is designed. The follow-up work in \cite{STCNet2019} conducte clustering on different Point of Interest (POI) modes, and then uses transfer learning in different communication traffic patterns to improve the prediction accuracy.
GCN-based methods \cite{wu2019graph, li2023dynamic, yao2021mvstgn} have been developed for WT forecasting on graph data.
Besides, DiffTAD \cite{li2024difftad} is a framework designed to detect anomalies in vehicle trajectories using advanced diffusion models. It aims to develop deep learning models that learn the reverse of the diffusion process, enabling the detection of anomalies by comparing the differences between a query trajectory and its reconstructed counterpart.
The aforementioned methods solely concentrate on the forms of aggregating  short-term input series, overlooking the challenges posed by the long-term and large-scale irregular WT data.

\subsection{Long-Term Time Series Forecasting}

When the sampled subseries becomes tractable, LL-WTF can be regarded as a typical long-term series forecasting task. 
There are several methods that can be adopted, including attention-based long-term forecasting \cite{Zhou2021Informer, wu2021autoformer, zhou2022fedformer, nie2022time, liang2023does, liang2024minusformer}.
For example, several works have improved the series aggregation forms of attention mechanism, such as operations of exponential intervals adopted in LogTrans \cite{li2019enhancing}, KL-divergence based ProbSparse activations in Informer \cite{Zhou2021Informer}, frequency-based random sampling in FEDformer \cite{zhou2022fedformer}, channel-independent patches in PatchTST \cite{nie2022time}, and period-based subseries integration in Periodformer \cite{liang2023does}. 
In summary, existing long-term time series forecasting methods have primarily focused on efficiently aggregating similar subseries by designing novel modules, rather than offering effective solutions for the crucial issue of non-stationarity existing in time series, making them relatively inefficient.




\subsection{Decomposition for Time Series Forecasting}


Time series exhibit a variety of patterns, and it is meaningful and beneficial to decompose them into several components, each representing an underlying category of patterns that evolving over time \cite{1976_timeseries}.
Several methods, e.g., STL \cite{cleveland1990stl} and Prophet \cite{taylor2018forecasting}, commonly utilize decomposition as a preprocessing phase on historical series.
There are also some methods, e.g., Autoformer \cite{wu2021autoformer}, FEDformer \cite{zhou2022fedformer} and Non-stationary Transformers \cite{liu2022non}, that harnesse decomposition into the attention module.
The authors in \cite{STWave+, STWave} present novel disentangle-fusion frameworks, STWave and STWave+, designed to address the issue of distribution shift. 
Furthermore, STHMLP is present in \cite{qin2023spatio} as a decomposition architecture to empower the model to capture multi-scale temporal dependencies.
The aforementioned methods attempt to apply decomposition to input series to enhance predictability, reduce computational complexity, or ameliorate the adverse effects of distribution shift.
However, these methods still rely on the stability assumption of the decomposed components, making it difficult to establish robust correlations with future series. 
In this paper, the proposed method progressively guides the learning of each time-varying pattern by decomposing the supervision signals to cope with the non-stationary problem. 
While models such as STWave, STWave+, and STHMLP utilize advanced decomposition techniques to capture spatio-temporal dependencies, our method distinguishes itself by progressively guiding the learning process. This comparison underscores the advantages of our approach in terms of backbone architecture, generalization capabilities, computational efficiency, and interpretability.

\section{Conclusion}
\label{sec_con}

In this paper, we focus on the task of long-term and large-scale wireless traffic forecasting (LL-WTF). 
To tackle the challenges posed by large-scale unstructured graph data and the long-term highly non-stationary nature of LL-WTF, we propose a Random Subgraph Sampling (RSS) algorithm. RSS repeatedly selects random subgraphs to facilitate the large-scale graph sampling. To solve the non-stationarity of Wireless Traffic (WT) data, we propose a Progressive Supervised learning paradigm based on Label Decomposition (PSLD). PSLD decomposes the label data into multiple easy-to-learn components, guides the deep learning model to learn meaningful component signals gradually in the shallow layer, and finally combines them into the final prediction results in the deep layer. The experimental results on three large-scale real WT datasets show that our method outperforms other methods in terms of prediction accuracy and runtime. PSLD effectively captures the non-stationarity of WT data, enhancing prediction accuracy while maintaining low computational cost and improved interpretability.
In future work, we will focus on incorporating more advanced non-linear models, such as Transformers and large language models, to more effectively capture the complexities of real-world interactions.


\bibliographystyle{IEEEtran}
\bibliography{citation}

\end{document}